\pdfoutput=1
\documentclass{article}

\PassOptionsToPackage{numbers, compress}{natbib}

\usepackage[main, final]{neurips_2026}

\usepackage[utf8]{inputenc}
\usepackage[T1]{fontenc}
\usepackage{hyperref}
\usepackage{url}
\usepackage{booktabs}
\usepackage{amsfonts}
\usepackage{nicefrac}
\usepackage{microtype}
\usepackage[table]{xcolor}
\usepackage{siunitx}
\usepackage{multirow}
\usepackage{amsmath}
\usepackage{cleveref}
\usepackage{graphicx}
\usepackage{pifont}
\usepackage{wrapfig}
\newcommand{\cmark}{\textcolor{green!60!black}{\ding{51}}}
\newcommand{\xmark}{\textcolor{red!60}{\ding{55}}}
\usepackage{caption}

\title{\textsc{Aeon}: Unifying Video and 3D World Models} 

\author{Weiqi Zhang\quad\quad\quad\;\; Wenyuan Zhang\quad\quad\quad\;\;  Yu-Shen Liu$^{\dagger}$\quad\quad\quad\;\; Junsheng Zhou$^{\dagger}$
\vspace{0.15cm}
\\School of Software, Tsinghua University, Beijing, China 
\vspace{0.15cm}
\\
{\tt\small \{zwq23, zhou-js24\}@mails.tsinghua.edu.cn\hspace{3mm} 
zhangweny98@gmail.com}
\\
{\tt\small liuyushen@tsinghua.edu.cn
}}
\begin{document}

\maketitle

\begingroup
\renewcommand{\thefootnote}{\fnsymbol{footnote}}
\footnotetext[2]{The corresponding authors are Yu-Shen Liu and Junsheng Zhou.}
\endgroup

{
    \centering
    \captionsetup{type=figure}
    \includegraphics[width=\textwidth]{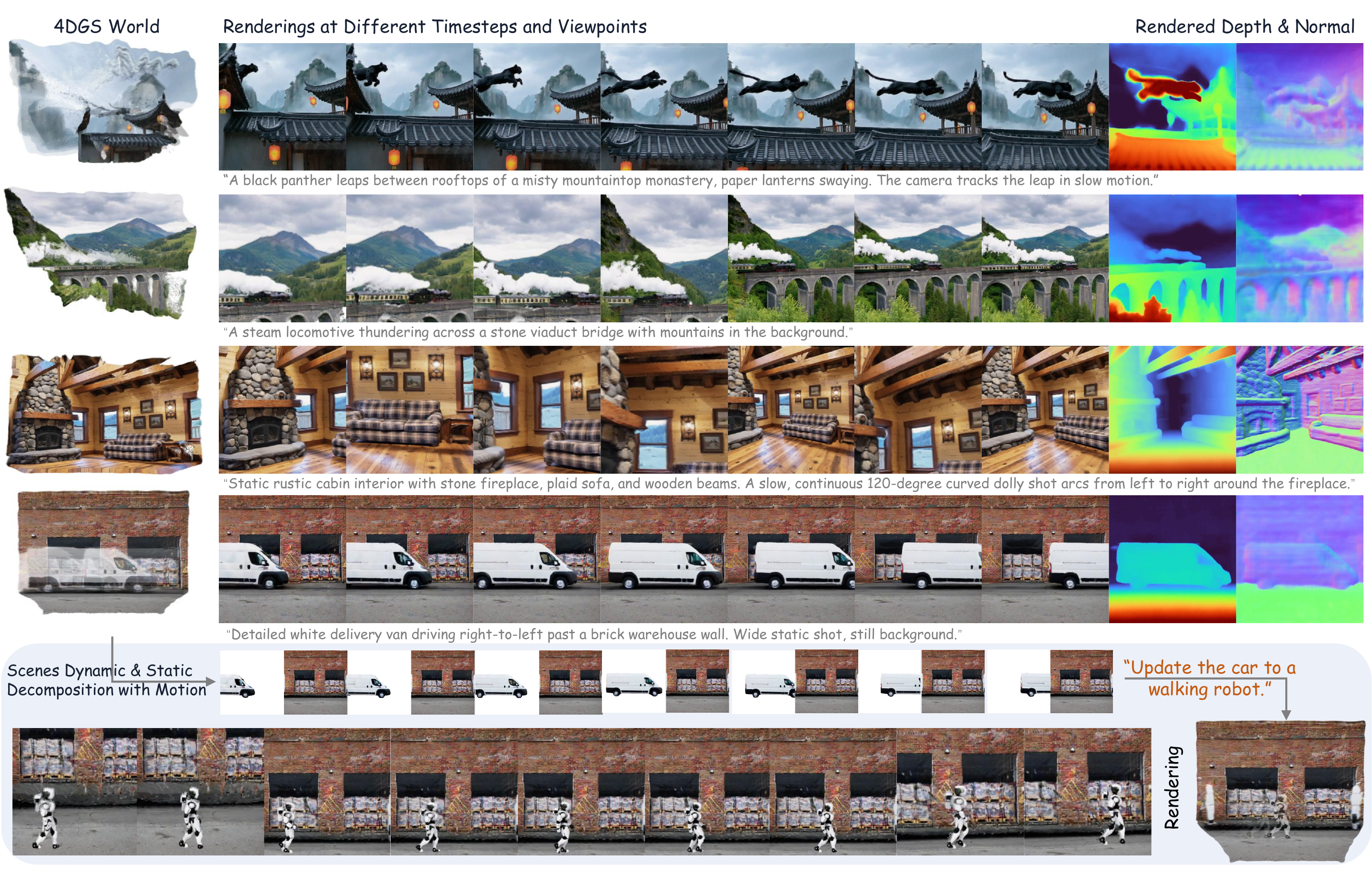}
    \captionof{figure}{\textsc{Aeon} generates 4DGS worlds that support real-time rendering at different timesteps and viewpoints. The 4DGS worlds are also ready for explicit editing and simulation (bottom).}
    \label{fig:teaser}
}

\begin{abstract}
\label{sec:abstract}
  Current world models for explicit world simulation typically follow two distinct pathways: (1) video generative models (e.g., Sora), which simulate realistic appearances but fail to preserve 3D consistency; and (2) 3D generative models (e.g., Marble), which generate geometrically consistent scenes but often yield static, less realistic observations. We present AEON, a unified framework that bridges video-based and 3D-based world models and combines the benefits of both. The key insight of AEON lies in a generative model that operates within the text- or image-conditioned video latent space while producing renderable 4D graphics primitives—specifically, 4D Gaussian Splatting (4DGS)—as the explicit world representation. We train a 4D reconstruction model to predict both 4DGS and camera poses from monocular videos, serving as our initialized 4D decoder. Furthermore, we design a domain adapter that translates generative video latents into reconstructive 4D latents through a distillation process on the video VAE. AEON naturally supports video-based world modeling by rendering 4DGS at predicted camera trajectories, and constructs a persistent 4D environment with scene dynamics for realistic world simulation. Consequently, AEON empowers a diverse range of world modeling applications and achieves state-of-the-art performance on standard benchmarks. \textit{Project Page: \url{https://weiqi-zhang.github.io/AEON}}.
\end{abstract}
\section{Introduction}
\label{sec:introduction}

World modeling has rapidly advanced to the forefront of AI research, driven by its immense potential in embodied AI, autonomous driving, and game design. Despite this progress, there remains a lack of consensus on the optimal representation for explicit world simulation, leaving the field divided between two dominant paradigms: video-based~\cite{openaisora2024,wan2025wan} and 3D-based world models~\cite{worldlabsmarble,hy2026hy}. As highlighted in Table~\ref{tab:comparison}, these approaches possess orthogonal strengths and suffer from fundamental, paradigm-specific bottlenecks. Consequently, a more powerful and unified representation is imperative to achieve comprehensive world modeling.

We begin by asking a fundamental question: What constitutes a good world model? To train robots to understand physical laws and perceive scenes, a world model should exhibit the following properties: (1) generating \textbf{realistic} appearances that align with real-world domains; (2) modeling \textbf{dynamic} scene evolution (e.g., fire, waves) and moving objects; (3) maintaining \textbf{3D consistent} environments free from geometric ambiguity; (4) ensuring scene \textbf{persistence} for continuous and repeatable simulation; and (5) supporting \textbf{real-time exploration} to facilitate efficient interaction. We argue that a representation unifying video- and 3D-based world models, thereby benefiting from their complementary strengths, is the key to achieving such an ideal world model.

\suppressfloats[t]
\begin{table}[t]
\centering
\caption{Comparison of different world modeling paradigms across fundamental characteristics.}
\label{tab:comparison}
\renewcommand{\arraystretch}{1.25}
\setlength{\tabcolsep}{1em}
\resizebox{\columnwidth}{!}{
\begin{tabular}{@{}lccccc@{}}
\toprule
\textbf{Paradigm} & \textbf{Realistic} & \textbf{3D Consistent} & \textbf{Dynamic} & \textbf{Persistence} & \textbf{Real-time Exp.} \\
\midrule
Video WM (e.g. Sora) & \cmark & \xmark & \cmark & \xmark & \xmark \\
3D WM (e.g. Marble)    & \xmark & \cmark & \xmark & \cmark & \cmark \\
\midrule
\textbf{AEON} & \cmark & \cmark & \cmark & \cmark & \cmark \\
\bottomrule
\end{tabular}}
\vspace{-0.3cm}
\end{table}

We present AEON, a generative world model that introduces 4D Gaussian Splatting (4DGS)~\cite{wang2025freetimegs,yangreal} as its foundational representation, unifying video and 3D models to benefit from both paradigms. 4DGS represents scenes as a set of explicit graphics primitives localized in spatial geometry, and models scene dynamics through the velocity-driven movement and time-varying visibility of these primitives. Consequently, a learned 4DGS representation provides two independent yet crucial capabilities: it models world appearances through direct rendering along user-defined camera trajectories, and it establishes a persistent 4D environment that supports highly controllable 3D simulations. Additionally, it enables explicit modeling of world geometry and physical states, including depth, normals, motion, and optical flow.

Despite the comprehensive capabilities inherent to the 4DGS representation, most existing approaches restrict its application to reconstruction tasks, and generating a persistent 4DGS world directly from text or image conditions remains largely unexplored. In AEON, we propose a novel framework that integrates the generative latent space of a pretrained video Diffusion Transformer (DiT)~\cite{wan2025wan,peebles2023scalable} with the reconstructive latent space of a 4D reconstruction model. Specifically, we train a 4D Gaussian Reconstruction Model (4DRM) to jointly predict 4DGS attributes and camera poses from monocular videos. By utilizing a large-scale annotated 4D video dataset and pretrained geometric priors from MapAnything~\cite{keetha2025mapanything}, a scalable geometry model, the 4DRM develops robust reconstructive capabilities to recover world geometry and dynamics. A primary challenge, however, is the significant domain gap between the video and 4DRM latent spaces. We address this by introducing a domain adapter that seamlessly bridges the two. Finally, we employ a multi-stage training pipeline that optimizes the adapter and fine-tunes the reconstruction model through end-to-end distillation from the video model, supervised by video VAE-decoded outputs.

Consequently, AEON unlocks an extensive array of world modeling applications, including text/image/video-to-4DGS generation, 3D world generation, robotics simulation, physical state estimation (e.g., depth, normals, and optical flow), scene editing, and novel view and timestep synthesis. 

Our main contributions can be summarized as follows:

\begin{itemize}
    \item We introduce AEON, a unified world modeling framework that represents worlds as explicit 4D Gaussians, effectively bridging video-based and 3D-based models to benefit from their complementary strengths.
    \item We propose a novel methodology that integrates the generative latent space of a pretrained video DiT with the reconstructive latent space of a 4D Gaussian reconstruction model. To achieve this, we train a specialized domain adapter through an end-to-end distillation of the video model.
    \item AEON demonstrates comprehensive versatility, natively supporting a diverse spectrum of world modeling tasks, and establishes new state-of-the-art results on standard benchmarks.
\end{itemize}

\section{Related Work}

\subsection{Video and 3D World Models}

World modeling aims to generate explorable and realistic environments from conditional inputs (e.g., text or images), serving as a foundation for embodied AI and autonomous systems. Current paradigms primarily rely on video-based~\cite{openaisora2024,veo22025,ren2025gen3c,kuaishou2024kling,cogvideox,wan2025wan} or 3D-based~\cite{worldlabsmarble,team2025hunyuanworld,yang2025matrix,hy2026hy,shen2026lyra,bahmani2026lyra,go2025vist3a} approaches. Within the video-centric domain, JEPA-style methods~\cite{assran2023self,assran2025v} excel at forecasting future latent states but lack explicit visual synthesis. Conversely, video world models~\cite{openaisora2024,wan2025wan,kuaishou2024kling} and their action-conditioned variants~\cite{bruce2024genie,wang2026matrix,team2026inspatio,team2026advancing} can synthesize photorealistic frames, yet they struggle with 3D consistency and incur heavy computational costs due to massive per-frame token generation.

Alternatively, 3D world models~\cite{worldlabsmarble,hy2026hy} construct explicit environments for simulation. Recent works have explored panoramic generation~\cite{team2025hunyuanworld,yang2025matrix}, progressive expansion~\cite{hy2026hy}, or leveraging video priors~\cite{bahmani2026lyra}. However, these methods typically employ static representations such as 3DGS~\cite{kerbl20233dgs} or meshes, precluding the modeling of world dynamics and limiting their ability to scale on dynamic internet video datasets. The same holds for 3D generation and reconstruction more broadly, which mostly build on static Gaussians~\cite{zhou2024diffgs,zhang2025gap,zhang2026gaussiangrow,zhang2025materialrefgs,xiang2026vggs} or distance fields~\cite{zhou2024udiff,zhou2026udfstudio,noda2025bijective,noda2024multipull}. To bridge this gap, we propose AEON, a framework that generates renderable 4D Gaussian Splatting~\cite{yangreal,wang2025freetimegs} (4DGS) primitives as an explicit world representation. AEON effectively unifies the strengths of both paradigms, achieving dynamic modeling with 3D consistency while overcoming their respective limitations.

\subsection{4D Gaussian Splatting from Videos}

4D Gaussian Splatting~\cite{yangreal,wang2025freetimegs} advances the success of 3D Gaussian Splatting~\cite{kerbl20233dgs} for modeling scene dynamics. 4DGS learns from multi-view videos to predict Gaussian primitives with explicit temporal attributes such as lifespan, velocity, and time-varying opacity, allowing Gaussians to appear, move, and vanish independently. However, predicting 4DGS from monocular videos remains significantly more challenging due to limited observational data. While optimization-based methods~\cite{wang2025shape,lei2025mosca,sun2024splatter} can achieve faithful visual appearances, they require lengthy fitting processes and struggle to generalize or scale. More recently, feed-forward frameworks~\cite{ren2024l4gm,xu20254dgt,yang2026neoverse,lin2026movies,zhang2026gamma} have been proposed for fast and scalable video-to-4DGS reconstruction, including bullet-time reconstruction at arbitrary timestamps~\cite{liang2025btimer} and online reconstruction from uncalibrated video streams~\cite{wu2026streamsplat}. Some of these works predict deformable 3DGS~\cite{lin2026movies,pan2025diff4splat} as their foundational representation, which often falls short when modeling complex and diverse dynamics. Related feed-forward models also recover dynamic geometry and camera poses from monocular videos~\cite{fang2026more}. All of these methods reconstruct scenes from observed videos. Native 4DGS generation has also been explored for animated objects~\cite{zhang2026tesla}. Closer to our setting, Diff4Splat~\cite{pan2025diff4splat} and Lyra~\cite{bahmani2026lyra} build Gaussian scenes on top of video diffusion priors, but both require a camera trajectory as input and represent dynamics with deformable 3DGS and per-frame 3DGS, respectively. AEON differs in both respects: it generates a world from a text prompt or a single image without camera input, and it represents the world as one persistent set of 4D primitives, each with its own reference time, duration, and velocity, so that content can appear, move, and vanish independently and can be rendered at any continuous timestep. Concurrent with our work, Beyond Pixels~\cite{liu2026beyond} also bypasses RGB decoding, lifting video latents to point maps instead of 4DGS.

\label{sec:relat}

\section{Method}
\label{sec:method}

\begin{figure}[t!]
    \centering
    \includegraphics[width=\linewidth]{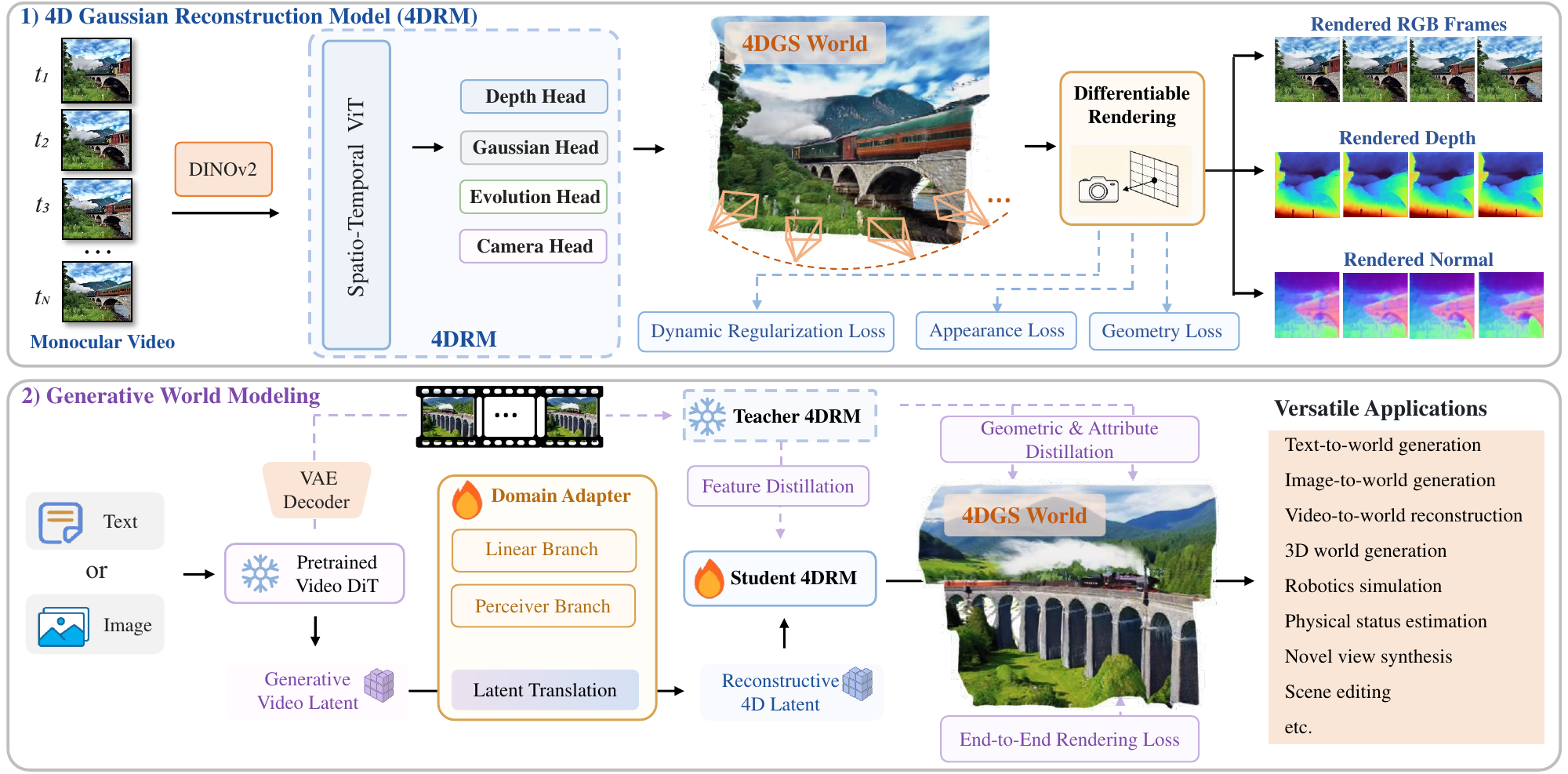}
    \caption{\textbf{Overview of AEON.}
    \textbf{Top (1).} The 4D Reconstruction Model (4DRM) decodes a monocular video into a 4DGS world through four task-specific heads that respectively predict depth, Gaussian appearance, dynamic evolution, and camera poses. The resulting primitives are rasterized into RGB, depth, and normal maps for end-to-end supervision.
    \textbf{Bottom (2).} We train AEON for generative world modeling by distilling the world generative capability of a frozen video diffusion model (e.g., Wan~2.2~\cite{wan2025wan}) and the 4D reconstructive capability of a teacher 4DRM via a Domain Adapter and a learnable student 4DRM. The generated explicit 4DGS world enables versatile world modeling applications.}
    \label{fig:pipeline}
\end{figure}

We present AEON, a generative world model that operates within the latent space of a pretrained video diffusion model while producing explicit 4D Gaussian Splatting (4DGS) as the world representation. Given a text prompt or conditioning image, our goal is to generate a set of 4D Gaussians encoding both appearance and dynamics, enabling photorealistic rendering and persistent 3D simulation. The overview of AEON is shown in Fig.~\ref{fig:pipeline}. We begin by introducing the 4DGS representation in Sec.~\ref{sec:3.1}, which models dynamic scenes through velocity-driven motion and time-varying visibility. In Sec.~\ref{sec:3.2}, we present the 4D Gaussian Reconstruction Model, whose four task-specific heads jointly predict depth, Gaussian appearance, dynamic evolution, and camera poses from monocular videos. We then describe our domain adapter in Sec.~\ref{sec:3.3}, which bridges the generative video latent space and the reconstructive 4D latent space through distillation. Finally, Sec.~\ref{sec:3.4} details our multi-stage training strategy that optimizes the full pipeline end-to-end.

\subsection{4D Gaussian Splatting as the World Representation}
\label{sec:3.1}
An ideal world model should jointly capture realistic appearances, scene dynamics, and 3D-consistent geometry within a persistent and efficiently renderable representation. To this end, AEON adopts 4D Gaussian primitives~\cite{wang2025freetimegs,yangreal,zhang2026gamma} as its foundational scene representation, where each primitive simultaneously encodes geometry, appearance, and temporal evolution. Formally, a single primitive $g_i$ is characterized by the following parameter set
\begin{equation}
g_i = \{ \mathbf{x}_i, \mathbf{s}_i, \mathbf{r}_i, o_i, \mathbf{c}_i, t_i, l_i, \mathbf{v}_i, \boldsymbol{\omega}_i \}.
\end{equation}
Specifically, $\mathbf{x}_i \in \mathbb{R}^3$ denotes the spatial center, $\mathbf{s}_i \in \mathbb{R}^2$ parameterizes anisotropic scaling along the principal axes, and $\mathbf{r}_i \in \mathbb{R}^4$ encodes the orientation as a unit quaternion. The base opacity $o_i \in [0,1]$ together with the color $\mathbf{c}_i \in \mathbb{R}^{3}$ jointly determine the visual appearance of each primitive.
 
In addition to these spatial attributes, four temporal parameters govern the dynamic behavior of each primitive. The reference timestep $t_i \in [0, T]$ specifies the moment of peak activation, while the duration $l_i \in \mathbb{R}^+$ controls the temporal extent over which the primitive remains visible. The linear velocity $\mathbf{v}_i \in \mathbb{R}^3$ and the angular velocity $\boldsymbol{\omega}_i \in \mathbb{R}^3$ in axis-angle form capture translational and rotational motion, respectively. This decomposition into explicit spatial and temporal components enables AEON to model both persistent scene structure and transient dynamic phenomena within a unified representation.
 
\subsection{4D Gaussian Reconstruction Model}
\label{sec:3.2}

While 4DGS offers an ideal world representation, directly generating 4D Gaussians from diverse conditions is highly challenging without proper guidance and geometric initialization. To address this, we first train a 4D Gaussian Reconstruction Model (4DRM) as a foundation, focusing on the more tractable task of world recovery from monocular videos. In a single feed-forward pass, 4DRM uses four task-specific heads to jointly predict depth, camera poses, and 4D Gaussian attributes of appearance and evolution. The model is trained end-to-end via differentiable rendering. Once trained, 4DRM facilitates AEON’s generative world modeling in two ways: (1) it provides a reconstructive foundation to initialize the student 4DRM that recovers the world from video latents, and (2) it acts as a teacher model, guiding the student to acquire 4D Gaussian modeling capabilities through distillation.

\paragraph{Encoding.}
Given an input video $V = \{I_t\}_{t=1}^{T}$ of $T$ frames, we extract per-frame dense features $\mathbf{F}_t \in \mathbb{R}^{H/p \times W/p \times D}$ using a pretrained DINOv2 encoder~\cite{oquab2023dinov2}. To inject temporal identity, we add a sinusoidal positional embedding of the timestep index $t$ to each spatial feature map. The resulting per-frame tokens are concatenated across all frames and processed by a transformer backbone of $L$ layers that alternate between intra-frame and cross-frame self-attention, balancing computational efficiency with long-range temporal modeling. The output features $\mathbf{F}^S \in \mathbb{R}^{THW/p^2 \times D}$ encode holistic spatio-temporal context for subsequent decoding. We initialize 4DRM from the pretrained weights of MapAnything~\cite{keetha2025mapanything}, a feed-forward metric 3D reconstruction model.

\paragraph{Decoding.}
We decode the shared spatio-temporal representation $\mathbf{F}^S_t$ through four task-specific heads, each responsible for a complementary component of the 4DGS world. The \emph{Depth Head} predicts dense per-frame depth maps $\mathbf{D}_t \in \mathbb{R}^{H \times W}$, while the Camera Head regresses the intrinsics $\mathbf{K}_t$ together with per-frame rotations $\mathbf{R}_t$ (unit quaternions) and translations $\mathbf{t}_t$. The predicted depths are unprojected via $\mathbf{K}_t$ into the Gaussian centers $\mathbf{x}_i$, anchoring all downstream predictions in a metrically grounded 3D space. Conditioned on these centers, the \emph{Gaussian Head} predicts the appearance attributes $\{\mathbf{s}_i, \mathbf{r}_i, o_i, \mathbf{c}_i\}$ (scale, orientation, opacity, and color), and the \emph{Evolution Head} predicts the dynamic attributes $\{t_i, l_i, \mathbf{v}_i, \boldsymbol{\omega}_i\}$ (temporal reference, duration, and linear and angular velocities) that govern how each primitive moves and fades over time. This decomposition lets the Gaussian and Evolution Heads focus exclusively on appearance and dynamics, while the Camera Head's joint pose estimation enables 4DRM to operate directly on raw monocular videos, removing the dependency on pre-estimated poses or external Structure-from-Motion (SfM) pipelines required by prior work~\cite{xu20254dgt, lin2026movies}. As a result, 4DRM recovers a complete dynamic world state from a single monocular video and establishes a robust reconstructive latent space for the subsequent generative integration.

\subsection{Generative World Modeling}
\label{sec:3.3}

\paragraph{Overview.}
AEON is designed to generate persistent and explorable 4D worlds from diverse conditions, such as text prompts and images. The complete pipeline is illustrated in the lower portion of Fig.~\ref{fig:pipeline}. Given a condition (e.g., a text prompt), we leverage a pretrained video diffusion model (e.g., Wan 2.2) as the generative foundation to produce a video latent, $z_{\mathrm{video}}$. Given that $z_{\mathrm{video}}$ encapsulates profound world-modeling priors capable of decoding into realistic, high-quality videos, we aim to unlock its full potential by recovering an explicit and persistent 4DGS world. We introduce a Domain Adapter that performs latent translation, mapping the generative video latent $z_{\mathrm{video}}$ to a reconstructive 4D latent $z_{\mathrm{4D}}$. A learnable student 4DRM then utilizes $z_{\mathrm{4D}}$ to explicitly recover the 4DGS world. A simpler alternative would decode $z_{\mathrm{video}}$ into frames with the VAE and pass them to the Stage-1 4DRM. We avoid this cascade because the 4DRM is trained on real videos, whereas decoded generations carry VAE and generation artifacts outside that distribution, so decoding errors would compound with reconstruction errors.

A primary challenge in connecting the video generative model to the 4D reconstruction model is the significant domain gap between their latent representations. The video latent is optimized for generative denoising, while the 4DRM consumes DINOv2~\cite{oquab2023dinov2} tokens that are trained for geometric and semantic understanding. These two feature spaces therefore occupy fundamentally different distributions. Naively replacing the DINOv2 patch embedding with a randomly initialized projection would forfeit the geometric prior carried by the pretrained backbone and force the student 4DRM to relearn its feature distribution from scratch. To avoid this, we introduce a learnable domain adapter that bridges the two latent spaces while preserving the geometric capabilities of the pretrained 4DRM backbone layers.
 
\paragraph{Dual-Path Domain Adapter.}
We implement the domain adapter as a dual-path architecture that couples a linearly initialized convolutional branch with an expressive Perceiver Bridge. To establish a robust baseline, we first solve a closed-form ridge regression to map the video latents to the target DINOv2 activations at each candidate insertion depth~$\ell$. This linear mapping already recovers a coarse alignment, which makes it a good initialization, although on its own it falls well short of the full adapter (Sec.~\ref{sec:ablation}). Motivated by this, our dual-path design employs a linear branch to provide a stable foundation, while an expressive residual branch captures complex nonlinear transformations. Specifically, the linear branch~$\mathbb{L}$ is instantiated as a strided 3D convolution initialized with the closed-form ridge solution. The residual branch~$\mathbb{N}$ comprises a Perceiver Bridge module, detailed below. A learnable scalar gate~$\gamma$ fuses the two branches as:
\begin{equation}
    Z = \mathbb{L}(z_{\mathrm{video}}) + \gamma \cdot \mathbb{N}(z_{\mathrm{video}}),
    \label{eq:adapter}
\end{equation}
where~$\gamma$ is initialized to zero. Consequently, the adapter's optimization begins entirely from the stable linear solution and progressively incorporates the nonlinear residual branch as~$\gamma$ updates via gradient descent, ensuring stable optimization from the outset.
 
\paragraph{Perceiver Bridge.}
Implementing the residual branch with standard convolutions inherently dilutes fine-grained geometric information due to their restricted local receptive fields. Instead, we design a highly expressive Perceiver~\cite{jaegle2021perceiver} Bridge module that leverages a learned token-to-token retrieval with a global receptive field. The video latent is first projected into the target feature dimension by a spatio-temporal Conv3D. An asymmetric temporal upsampler then restores the original temporal resolution by bypassing the independently encoded first frame and applying sub-pixel shuffle to the remaining compressed tokens so that the output length matches the original video exactly. The upsampled tokens are passed into a stack of Perceiver-style cross-attention blocks, where a set of learnable spatial queries attend to the projected video tokens as key-value pairs. Both queries and keys carry fixed 2D sinusoidal positional embeddings, enabling explicit spatial alignment without any resolution-specific learned parameters. Each frame is processed independently through the stack, confining the attention cost to the two-dimensional spatial grid while temporal modeling is handled by the preceding Conv3D and upsampler. A final LayerNorm aligns the output distribution with that of the DINOv2 tokens.
 
\paragraph{Token-Level Injection.}
The adapter output $Z$ is integrated into the DINOv2 backbone at layer $\ell$ by bypassing the patch embedding and the initial $\ell$ transformer blocks, utilizing $Z$ as the input tokens for the remaining blocks. Notably, these subsequent DINOv2 transformer blocks are fused into the 4DRM as part of the learnable spatio-temporal ViT. The selection of $\ell$ requires a careful trade-off: integrating the features too early imposes an excessive optimization burden on the adapter, whereas a late integration sacrifices essential geometric features relied upon by the downstream heads. Therefore, we set $\ell$ to the shallowest layer where the linear branch yields a reliable initialization. This strategy maximizes the number of retained upper blocks, effectively incorporating DINOv2's geometric knowledge into the 4DRM.

We train the student 4DRM and the domain adapter with both distillation losses and end-to-end rendering supervisions. The full training procedure is detailed in Sec.~\ref{sec:3.4}.

\subsection{Training Strategy}
\label{sec:3.4}

We adopt a multi-stage pipeline that first builds a strong reconstructive backbone and then bridges the generative video latent space through generative distillation and end-to-end rendering refinement.

\textbf{Stage 1: Reconstruction Model Training.}
The 4DRM learns to map monocular video to 4D Gaussian primitives through large-scale training on posed videos. From each training clip of $T$ frames, we select $K$ frames as network input and predict a unified set of 4D Gaussians. These Gaussians are rendered back to all $T$ frames via the differentiable rasterizer, yielding dense photometric supervision that extends well beyond the temporal coverage of the input. Our objective consists of three complementary terms that respectively enforce appearance fidelity, geometric grounding, and dynamic disentanglement.

\noindent\emph{Appearance fidelity.} We supervise rendered frames $\tilde{\mathbf{I}}_t$ against ground-truth observations $\mathbf{I}_t$ with a composite loss that balances pixel accuracy and perceptual quality:
\begin{equation}
    \mathcal{L}_{\text{app}} = \frac{1}{T}\sum_{t=1}^{T} \left( \|\mathbf{I}_t - \tilde{\mathbf{I}}_t\|_2^2 + \alpha \cdot \text{LPIPS}(\mathbf{I}_t, \tilde{\mathbf{I}}_t) + \beta \cdot \left( 1 - \text{SSIM}(\mathbf{I}_t, \tilde{\mathbf{I}}_t) \right) \right).
\end{equation}
The perceptual and structural terms prevent convergence to overly smooth solutions and encourage sharp, realistic textures in the rendered output.

\noindent\emph{Geometric grounding.} Monocular video alone cannot resolve metric depth ambiguity. We therefore supervise predicted geometry directly with ground-truth depth and normal annotations from our training dataset. Specifically, we render per-frame depth $\tilde{\mathbf{d}}_t$ and normals $\tilde{\mathbf{n}}_t$ from the predicted 4DGS and minimize their discrepancy with the annotated counterparts $\mathbf{d}_t^*$ and $\mathbf{n}_t^*$:
\begin{equation}
    \mathcal{L}_{\text{geo}} = \frac{1}{T}\sum_{t=1}^{T} \left( \mu \|\mathbf{d}_t^* - \tilde{\mathbf{d}}_t\|_2^2 + \nu \|\mathbf{n}_t^* - \tilde{\mathbf{n}}_t\|_2^2 \right).
\end{equation}
Unlike prior methods that rely on pseudo labels from monocular expert models, training against metric ground truth provides an unambiguous supervisory signal and anchors all predictions in a consistent world-scale coordinate frame.

\noindent\emph{Dynamic disentanglement.} Without multi-view constraints, the network may collapse to a degenerate solution where every primitive is assigned a short lifespan, leading the representation to per-frame point clouds. To encourage proper separation of static structure from dynamic content, we penalize unnecessary motion and favor long-lived primitives:
\begin{equation}
    \mathcal{L}_{\text{dyn}} = \frac{1}{M}\sum_{i=1}^{M} \left( \eta_1 \|\mathbf{v}_i\|_1 + \eta_2 \|\boldsymbol{\omega}_i\|_1 + \eta_3 \, l_i^{-1} \right),
\end{equation}
where $M$ is the total number of predicted Gaussians. These penalties ensure that non-trivial velocity and short lifespans are assigned only to genuinely dynamic elements, enabling the model to learn meaningful motion decomposition without explicit motion labels.

The complete Stage 1 objective combines all three components:
\begin{equation}
    \mathcal{L}_{\text{s1}} = \mathcal{L}_{\text{app}} + \mathcal{L}_{\text{geo}} + \mathcal{L}_{\text{dyn}}.
\end{equation}

\textbf{Stage 2: Generative Distillation.}
In the second stage, we connect the generative video latent space to the reconstructive 4D space through end-to-end distillation. We freeze the pretrained 4DRM as a teacher and train the domain adapter alongside a student 4DRM. The teacher takes ground-truth RGB frames as input, while the domain adapter and student 4DRM operate solely on video latents. The training objective drives the student to reproduce the teacher's 4DGS world reconstruction in a generative manner from the video latents. The overall distillation loss combines three complementary terms:
\begin{equation}
\mathcal{L}_{\text{distill}} = \mathcal{L}_{\text{4DGS}} + \mathcal{L}_{\text{geom}} + \lambda(t)\,\mathcal{L}_{\text{feat}}.
\end{equation}

\textit{4D Gaussian attribute alignment.} $\mathcal{L}_{\text{4DGS}}$ applies a weighted L1 loss between the per-primitive attributes predicted by the teacher and the student, covering spatial position, anisotropic scale, rotation, opacity, color, velocity, and lifespan. This term provides the most direct form of knowledge transfer by enforcing primitive-level correspondence in attribute space.

\textit{Dense geometric alignment.} Attribute-space supervision alone is insufficient since distinct sets of Gaussian attributes can produce visually equivalent renderings, while small attribute deviations may cause large visual discrepancies. To address this ambiguity, $\mathcal{L}_{\text{geom}}$ supervises the per-view dense geometric predictions from the student 4DRM heads, combining a robust log-depth regression loss, a multi-scale depth gradient term for boundary sharpness, and an L1 penalty on predicted camera parameters. We apply a global scale normalization before computing this loss to eliminate the trivial scale ambiguity between the two models.

\textit{Feature alignment.} $\mathcal{L}_{\text{feat}}$ regularizes the domain adapter by aligning the student's intermediate features at the bridge output with the corresponding teacher activations through an L1 loss. Its weight $\lambda(t)$ follows a cosine decay schedule that imposes strong regularization in early training to prevent catastrophic drift of the adapter and gradually relaxes to let the task-specific losses dominate.

Together, these objectives supervise the distillation process at complementary granularities. The attribute loss effectively instills the teacher’s 4D reconstruction capabilities into the student, while the geometric and feature losses simultaneously drive domain adaptation by forcing the adapter to project the video latent space onto the reconstructive 4D latent space. 

\textbf{Stage 3: End-to-End Rendering Refinement.} Distillation losses operate in attribute and feature space and do not directly reflect final rendering quality. We therefore fine-tune the domain adapter and student 4DRM with an end-to-end rendering loss, where pseudo ground-truth frames are obtained by decoding $z_{\mathrm{video}}$ through the frozen VAE decoder and compared against the rasterized 4DGS output using the same appearance objective as Stage~1. Gradients flow through the differentiable rasterizer into the full pipeline, so the student is optimized for rendering quality directly rather than only imitating the teacher. A linearly decayed distillation regularizer preserves the geometric structure acquired in Stage~2.

\section{Experiments}
\label{exper}

\begin{figure}[htbp]
    \centering
    \includegraphics[width=\linewidth]{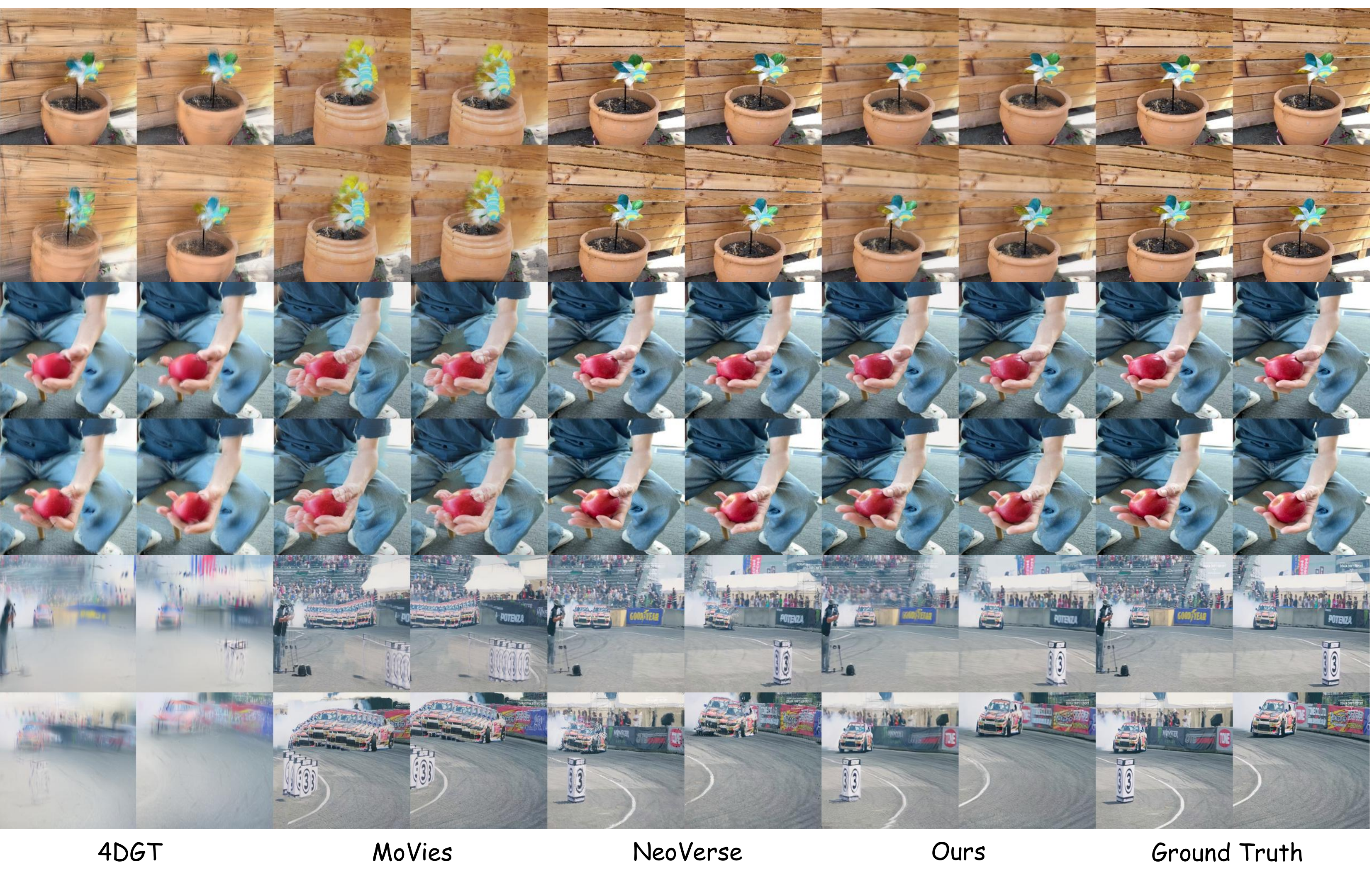}
    \caption{Qualitative reconstruction comparison on Dycheck and DaVis. }
    \label{fig:recon_qual}
    \vspace{-0.5cm}
\end{figure}

\subsection{4D Generation}

We evaluate AEON against 4DNex~\cite{chen20254dnex} and Lyra~\cite{bahmani2026lyra} for 4D generation. Unlike 4DNex (per-frame point clouds) and Lyra (per-frame 3DGS), AEON predicts a unified 4DGS representation for real-time exploration. We evaluate on 30 scenes, each specified by an image-text prompt. Evaluation videos, rendered along shared camera trajectories from identical image-text prompts, are assessed via VBench~\cite{huang2023vbench} metrics: I2V Subject, Subject Consistency, Motion Smoothness, and Aesthetic Quality. As shown in Table~\ref{tab:4d_gen}, AEON outperforms all baselines. While per-frame reconstruction lacks cross-frame coherence, AEON models scenes via persistent 4D Gaussians with explicit velocity and duration, naturally ensuring temporal consistency. Additionally, AEON generalizes across backbones (Wan~2.1 and Wan~2.2) without architectural modifications, yielding consistently strong results.

\begin{table}[htbp]
\centering
\caption{Quantitative comparison on 4D generation. All outputs are rendered along shared preset camera trajectories and scored with VBench~\cite{huang2023vbench} metrics.}
\label{tab:4d_gen}
\resizebox{\linewidth}{!}{
\begin{tabular}{lcccc}
\toprule
Method &
\textbf{I2V\_Subject $\uparrow$} &
\textbf{Subject\_Consistency $\uparrow$} &
\textbf{Motion\_Smoothness $\uparrow$} &
\textbf{Aesthetic\_Quality $\uparrow$} \\
\midrule
4DNex~\cite{chen20254dnex}  & 0.796 & 0.790 & 0.968 & 0.342 \\
Lyra~\cite{bahmani2026lyra}   & 0.865   & 0.864   & 0.974   & 0.459   \\
\midrule
AEON (Wan 2.1) & 0.912   & 0.864   & 0.976   & 0.519   \\
AEON (Wan 2.2) & \textbf{0.913} & \textbf{0.875} & \textbf{0.980} & \textbf{0.531} \\
\bottomrule
\end{tabular}
}
\end{table}

\subsection{Video-to-4D Reconstruction}

We evaluate AEON for 4D reconstruction on Dycheck~\cite{gao2022dynamic} and DaVis~\cite{Perazzi_CVPR_2016} against SOTA baselines: L4GM~\cite{ren2024l4gm}, 4DGT~\cite{xu20254dgt}, MoVies~\cite{lin2026movies}, NeoVerse~\cite{yang2026neoverse}, and StreamSplat~\cite{wu2026streamsplat}. Unlike 4DGT and MoVies, which rely on external poses (from GT or $\pi^3$~\cite{wang2025pi}), AEON’s integrated camera head jointly predicts camera parameters and 4DGS attributes from monocular RGB. StreamSplat additionally requires monocular depth from Depth Anything V2~\cite{yang2024depthv2}, which AEON does not need. For Dycheck’s novel view synthesis, we assess our video-to-4DGS pipeline by encoding input video into the VAE latent space and decoding it via the domain adapter and student reconstruction model.

\begin{wraptable}{r}{0.65\textwidth} 
    \centering
    \vspace{-0.2cm}
    \caption{Video-to-4D reconstruction comparison.}
    \vspace{-0.2cm}
    \label{tab:recon}
    \setlength{\tabcolsep}{6pt} 
    \resizebox{\linewidth}{!}{ 
        \begin{tabular}{l S[table-format=2.3] S[table-format=1.3] S[table-format=1.3] S[table-format=2.3] S[table-format=1.3] S[table-format=1.3]}
        \toprule
        \multirow{2}{*}{Method} & \multicolumn{3}{c}{\textbf{Dycheck}} & \multicolumn{3}{c}{\textbf{DaVis}} \\
        \cmidrule(lr){2-4} \cmidrule(lr){5-7}
        & {PSNR $\uparrow$} & {SSIM $\uparrow$} & {LPIPS $\downarrow$} & {PSNR $\uparrow$} & {SSIM $\uparrow$} & {LPIPS $\downarrow$} \\
        \midrule
        L4GM~\cite{ren2024l4gm} & 8.673 & 0.354 & 0.652 & 20.589 & 0.667 & 0.311 \\
        4DGT~\cite{xu20254dgt} & 12.267 & 0.413 & 0.542 & 20.362 & 0.667 & 0.313 \\
        MoVies~\cite{lin2026movies} & 11.840 & 0.341 & 0.545 & 20.736 & 0.688 & 0.307 \\
        NeoVerse~\cite{yang2026neoverse} & 14.269 & 0.363 & 0.501 & 23.929 & 0.735 & 0.283 \\
        StreamSplat~\cite{wu2026streamsplat} & 11.876 & 0.375 & 0.532 & 23.748 & 0.712 & 0.273 \\
        \midrule
        AEON & \textbf{14.487} & \textbf{0.421} & \textbf{0.469} & \textbf{24.661} & \textbf{0.771} & \textbf{0.243} \\
        \midrule
        \textit{Teacher 4DRM} & {\textit{14.413}} & {\textit{0.418}} & {\textit{0.462}} & {\textit{24.886}} & {\textit{0.775}} & {\textit{0.241}} \\
        \bottomrule
        \end{tabular}
    }
    \vspace{-0.4cm}
\end{wraptable}
As summarized in Table~\ref{tab:recon}, AEON achieves superior PSNR, SSIM, and LPIPS across both benchmarks. Despite operating through the video VAE latent bottleneck, AEON’s domain adapter effectively preserves the geometric details necessary for high-fidelity reconstruction, with especially pronounced gains on complex DaVis sequences. Furthermore, AEON needs neither camera poses nor depth at inference, and only Stage~1 uses annotated videos, so the later training stages can scale to raw videos. Qualitative comparisons in Fig.~\ref{fig:recon_qual} further corroborate these quantitative trends.

\subsection{Novel Trajectory Rendering}

We collect 30 in-the-wild videos unseen during training and render each scene along several novel camera trajectories. We compare against TrajectoryCrafter~\cite{Yu_2025_ICCV}, RecamMaster~\cite{bai2025recammaster}, and NeoVerse~\cite{yang2026neoverse}, reporting six VBench~\cite{huang2023vbench} metrics in Table~\ref{tab:novel_traj}. AEON achieves the best scores on four out of six metrics while remaining competitive on the other two. Since rendering a new trajectory from 4DGS is geometrically exact rasterization rather than learned hallucination, AEON naturally maintains strong consistency and visual quality across viewpoint changes.

\begin{table}[htbp]
\vspace{-0.3cm}
\centering
\setlength{\tabcolsep}{6pt} 
\caption{Novel trajectory rendering comparison under VBench~\cite{huang2023vbench} metrics.}
\label{tab:novel_traj}
\resizebox{\linewidth}{!}{
\begin{tabular}{lcccccc}
\toprule
Method & Subj. Consist. $\uparrow$ & Back. Consist. $\uparrow$ & Temp. Flick. $\uparrow$ & Motion Smooth. $\uparrow$ & Aesth. Quality $\uparrow$ & Imag. Quality $\uparrow$ \\
\midrule
TrajectoryCrafter~\cite{Yu_2025_ICCV} & 94.48 & 96.57 & 95.49 & \textbf{98.79} & 64.81 & 58.53 \\
RecamMaster~\cite{bai2025recammaster} & 94.49 & 95.58 & \textbf{97.64} & 97.79 & 61.26 & 56.34 \\
NeoVerse~\cite{yang2026neoverse} & 90.24 & 95.21 & 95.01 & 97.39 & 59.17 & 57.98 \\
\midrule
AEON & \textbf{94.64} & \textbf{96.77} & 96.69 & 98.11 & \textbf{69.02} & \textbf{58.65} \\
\bottomrule
\end{tabular}
}
\vspace{-0.5cm}
\end{table}

\subsection{Ablation Studies}
\label{sec:ablation}

\begin{wraptable}{r}{0.65\textwidth}
    \centering
    \vspace{-0.6cm}
    \caption{Ablation study on Dycheck and DaVis.}
    \label{tab:ablation}
    \setlength{\tabcolsep}{6pt}
    \resizebox{\linewidth}{!}{
        \begin{tabular}{lcccccc}
        \toprule
        \multirow{2}{*}{Variant}
        & \multicolumn{3}{c}{Dycheck}
        & \multicolumn{3}{c}{DaVis} \\
        \cmidrule(lr){2-4} \cmidrule(lr){5-7}
        & PSNR $\uparrow$ & SSIM $\uparrow$ & LPIPS $\downarrow$
        & PSNR $\uparrow$ & SSIM $\uparrow$ & LPIPS $\downarrow$ \\
        \midrule
        Conv3D projection & 12.086 & 0.314 & 0.606 & 12.813 & 0.276 & 0.701 \\
        W/o Distill  & 12.390 & 0.349 & 0.537 & 16.718 & 0.503 & 0.489 \\
        W/o Duration & 12.500 & 0.372 & 0.535 & 22.057 & 0.718 & 0.328 \\
        \midrule
        Full         & \textbf{14.487} & \textbf{0.421} & \textbf{0.469} & \textbf{24.661} & \textbf{0.771} & \textbf{0.243} \\
        \bottomrule
        \end{tabular}
    }
    \vspace{-0.2cm}
\end{wraptable}
We ablate three key design choices of AEON under the Dycheck and DaVis benchmarks in Table~\ref{tab:ablation}. Conv3D projection replaces the full dual-path domain adapter with a single Conv3D layer, showing that the expressive Perceiver Bridge branch is essential for capturing the complex mapping between the two latent spaces. \textit{W/o Distill} removes the distillation losses and trains the adapter with rendering supervision alone, resulting in a significant performance drop, especially on DaVis, which highlights the importance of the teacher 4DRM as a geometric foundation. \textit{W/o Duration} disables the temporal opacity evolution of 4DGS, forcing all primitives to remain visible throughout the sequence; the degradation shows that explicit duration modeling of 4DGS representation is critical for handling transient dynamics. The full model combines all components and achieves the best performance on every metric. 

\section{Conclusion}
\label{conclusion}

We present AEON, a unified world modeling framework that bridges video-based and 3D-based paradigms and combines the benefits of both. The core of AEON is a generative model that operates within the latent space of a pretrained video diffusion model while producing explicit 4D Gaussian Splatting as the world representation. We train a 4DRM for robust geometric guidance and design a domain adapter that translates generative video latents into the reconstructive 4D latents through end-to-end distillation. Experiments across 4D generation, dynamic-scene reconstruction, and novel trajectory rendering demonstrate that AEON achieves state-of-the-art performance while supporting a broad spectrum of world modeling applications within a single unified framework.

\paragraph{Limitations and future work.}
The temporal extent of generated 4D scenes is currently bounded by the video diffusion model's sequence length. Because AEON stores the world as a persistent 4DGS rather than as frames, it offers a natural path beyond this window: render conditioning frames from the current 4DGS at the latest camera pose, generate the next window, decode it into new primitives, and fuse them with the existing world where they overlap. Since each step is anchored to explicit geometry rather than extrapolated from previous frames, drift should accumulate more slowly than in video-only autoregression. In addition, regions never observed in the generated video cannot be recovered, so very large viewpoint changes may reveal holes that a generative completion module could fill.

\begin{ack}
This work was partially supported by Deep Earth Probe and Mineral Resources Exploration -- National Science and Technology Major Project (2024ZD1003405) and the National Natural Science Foundation of China (62272263).
\end{ack}

\bibliographystyle{ieee_fullname}
\bibliography{sample}

\begin{thebibliography}{10}\itemsep=-1pt

\bibitem{assran2025v}
Mido Assran, Adrien Bardes, David Fan, Quentin Garrido, Russell Howes, Mojtaba
  Komeili, Matthew Muckley, Ammar Rizvi, Claire Roberts, Koustuv Sinha, Artem
  Zholus, et~al.
\newblock {V-JEPA} 2: Self-supervised video models enable understanding,
  prediction and planning.
\newblock {\em arXiv preprint arXiv:2506.09985}, 2025.

\bibitem{assran2023self}
Mahmoud Assran, Quentin Duval, Ishan Misra, Piotr Bojanowski, Pascal Vincent,
  Michael Rabbat, Yann LeCun, and Nicolas Ballas.
\newblock Self-supervised learning from images with a joint-embedding
  predictive architecture.
\newblock In {\em Proceedings of the IEEE/CVF Conference on Computer Vision and
  Pattern Recognition (CVPR)}, 2023.

\bibitem{bahmani2026lyra}
Sherwin Bahmani, Tianchang Shen, Jiawei Ren, Jiahui Huang, Yifeng Jiang,
  Haithem Turki, Andrea Tagliasacchi, David~B. Lindell, Zan Gojcic, Sanja
  Fidler, Huan Ling, Jun Gao, and Xuanchi Ren.
\newblock Lyra: Generative {3D} scene reconstruction via video diffusion model
  self-distillation.
\newblock In {\em International Conference on Learning Representations (ICLR)},
  2026.

\bibitem{bai2025recammaster}
Jianhong Bai, Menghan Xia, Xiao Fu, Xintao Wang, Lianrui Mu, Jinwen Cao, Zuozhu
  Liu, Haoji Hu, Xiang Bai, Pengfei Wan, et~al.
\newblock {ReCamMaster}: Camera-controlled generative rendering from a single
  video.
\newblock In {\em Proceedings of the IEEE/CVF International Conference on
  Computer Vision (ICCV)}, 2025.

\bibitem{bruce2024genie}
Jake Bruce, Michael~D. Dennis, Ashley Edwards, Jack Parker-Holder, Yuge Shi,
  Edward Hughes, Matthew Lai, Aditi Mavalankar, Richie Steigerwald, Chris Apps,
  et~al.
\newblock Genie: Generative interactive environments.
\newblock In {\em International Conference on Machine Learning (ICML)}, 2024.

\bibitem{chen2024panda70m}
Tsai-Shien Chen, Aliaksandr Siarohin, Willi Menapace, Ekaterina Deyneka,
  Hsiang-wei Chao, Byung~Eun Jeon, Yuwei Fang, Hsin-Ying Lee, Jian Ren,
  Ming-Hsuan Yang, and Sergey Tulyakov.
\newblock {Panda-70M}: Captioning {70M} videos with multiple cross-modality
  teachers.
\newblock In {\em Proceedings of the IEEE/CVF Conference on Computer Vision and
  Pattern Recognition (CVPR)}, 2024.

\bibitem{chen20254dnex}
Zhaoxi Chen, Tianqi Liu, Long Zhuo, Jiawei Ren, Zeng Tao, He Zhu, Fangzhou
  Hong, Liang Pan, and Ziwei Liu.
\newblock {4DNeX}: Feed-forward {4D} generative modeling made easy.
\newblock {\em arXiv preprint arXiv:2508.13154}, 2025.

\bibitem{fang2026more}
Juntong Fang, Zequn Chen, Weiqi Zhang, Donglin Di, Xuancheng Zhang, Chengmin
  Yang, and Yu-Shen Liu.
\newblock {MoRe}: Motion-aware feed-forward {4D} reconstruction transformer.
\newblock In {\em Proceedings of the IEEE/CVF Conference on Computer Vision and
  Pattern Recognition (CVPR)}, 2026.

\bibitem{gao2022dynamic}
Hang Gao, Ruilong Li, Shubham Tulsiani, Bryan Russell, and Angjoo Kanazawa.
\newblock Monocular dynamic view synthesis: A reality check.
\newblock In {\em Advances in Neural Information Processing Systems (NeurIPS)},
  2022.

\bibitem{go2025vist3a}
Hyojun Go, Dominik Narnhofer, Goutam Bhat, Prune Truong, Federico Tombari, and
  Konrad Schindler.
\newblock Text-to-{3D} by stitching a multi-view reconstruction network to a
  video generator.
\newblock In {\em International Conference on Learning Representations (ICLR)},
  2026.

\bibitem{veo22025}
{Google DeepMind}.
\newblock {Veo 2}.
\newblock \url{https://deepmind.google/technologies/veo/veo-2/}, 2024.

\bibitem{HuCVPR2021}
Yuan-Ting Hu, Jiahong Wang, Raymond~A. Yeh, and Alexander~G. Schwing.
\newblock {SAIL-VOS 3D}: A synthetic dataset and baselines for object detection
  and {3D} mesh reconstruction from video data.
\newblock In {\em Proceedings of the IEEE/CVF Conference on Computer Vision and
  Pattern Recognition (CVPR)}, 2021.

\bibitem{huang2023vbench}
Ziqi Huang, Yinan He, Jiashuo Yu, Fan Zhang, Chenyang Si, Yuming Jiang, Yuanhan
  Zhang, Tianxing Wu, Qingyang Jin, Nattapol Chanpaisit, Yaohui Wang, Xinyuan
  Chen, Limin Wang, Dahua Lin, Yu Qiao, and Ziwei Liu.
\newblock {VBench}: Comprehensive benchmark suite for video generative models.
\newblock In {\em Proceedings of the IEEE/CVF Conference on Computer Vision and
  Pattern Recognition (CVPR)}, 2024.

\bibitem{hy2026hy}
Team HY-World, Chenjie Cao, Xuhui Zuo, Zhenwei Wang, Yisu Zhang, Junta Wu,
  Zhenyang Liu, Yuning Gong, Yang Liu, Bo Yuan, et~al.
\newblock {HY-World} 2.0: A multi-modal world model for reconstructing,
  generating, and simulating {3D} worlds.
\newblock {\em arXiv preprint arXiv:2604.14268}, 2026.

\bibitem{jaegle2021perceiver}
Andrew Jaegle, Felix Gimeno, Andy Brock, Oriol Vinyals, Andrew Zisserman, and
  Joao Carreira.
\newblock Perceiver: General perception with iterative attention.
\newblock In {\em International Conference on Machine Learning (ICML)}, 2021.

\bibitem{karaev2023dynamicstereo}
Nikita Karaev, Ignacio Rocco, Benjamin Graham, Natalia Neverova, Andrea
  Vedaldi, and Christian Rupprecht.
\newblock {DynamicStereo}: Consistent dynamic depth from stereo videos.
\newblock In {\em Proceedings of the IEEE/CVF Conference on Computer Vision and
  Pattern Recognition (CVPR)}, 2023.

\bibitem{keetha2025mapanything}
Nikhil Keetha, Norman M{\"u}ller, Johannes Sch{\"o}nberger, Lorenzo Porzi,
  Yuchen Zhang, Tobias Fischer, Arno Knapitsch, Duncan Zauss, Ethan Weber,
  Nelson Antunes, et~al.
\newblock {MapAnything}: Universal feed-forward metric {3D} reconstruction.
\newblock In {\em International Conference on 3D Vision (3DV)}, 2026.

\bibitem{kerbl20233dgs}
Bernhard Kerbl, Georgios Kopanas, Thomas Leimk{\"u}hler, and George Drettakis.
\newblock {3D} {Gaussian} splatting for real-time radiance field rendering.
\newblock {\em ACM Transactions on Graphics (TOG)}, 42(4):139:1--139:14, 2023.

\bibitem{kuaishou2024kling}
{Kuaishou}.
\newblock {Kling AI}.
\newblock \url{https://klingai.kuaishou.com/}, 2024.

\bibitem{lei2025mosca}
Jiahui Lei, Yijia Weng, Adam~W. Harley, Leonidas Guibas, and Kostas Daniilidis.
\newblock {MoSca}: Dynamic {Gaussian} fusion from casual videos via {4D} motion
  scaffolds.
\newblock In {\em Proceedings of the IEEE/CVF Conference on Computer Vision and
  Pattern Recognition (CVPR)}, 2025.

\bibitem{liang2025btimer}
Hanxue Liang, Jiawei Ren, Ashkan Mirzaei, Antonio Torralba, Ziwei Liu, Igor
  Gilitschenski, Sanja Fidler, Cengiz Oztireli, Huan Ling, Zan Gojcic, and
  Jiahui Huang.
\newblock Feed-forward bullet-time reconstruction of dynamic scenes from
  monocular videos.
\newblock In {\em Advances in Neural Information Processing Systems (NeurIPS)},
  2025.

\bibitem{lin2026movies}
Chenguo Lin, Yuchen Lin, Panwang Pan, Yifan Yu, Tao Hu, Honglei Yan, Katerina
  Fragkiadaki, and Yadong Mu.
\newblock {MoVieS}: Motion-aware {4D} dynamic view synthesis in one second.
\newblock In {\em Proceedings of the IEEE/CVF Conference on Computer Vision and
  Pattern Recognition (CVPR)}, 2026.

\bibitem{liu2026beyond}
Zihao Liu, Xiaolong Shen, Zhenglin Zhou, Ruijie Quan, and Yi Yang.
\newblock {Beyond Pixels}: From video priors to {4D} worlds.
\newblock {\em arXiv preprint arXiv:2608.10744}, 2026.

\bibitem{Mehl2023_Spring}
Lukas Mehl, Jenny Schmalfuss, Azin Jahedi, Yaroslava Nalivayko, and Andr\'es
  Bruhn.
\newblock Spring: A high-resolution high-detail dataset and benchmark for scene
  flow, optical flow and stereo.
\newblock In {\em Proceedings of the IEEE/CVF Conference on Computer Vision and
  Pattern Recognition (CVPR)}, 2023.

\bibitem{noda2024multipull}
Takeshi Noda, Chao Chen, Weiqi Zhang, Xinhai Liu, Yu-Shen Liu, and Zhizhong
  Han.
\newblock {MultiPull}: Detailing signed distance functions by pulling
  multi-level queries at multi-step.
\newblock In {\em Advances in Neural Information Processing Systems (NeurIPS)},
  2024.

\bibitem{noda2025bijective}
Takeshi Noda, Chao Chen, Junsheng Zhou, Weiqi Zhang, Yu-Shen Liu, and Zhizhong
  Han.
\newblock Learning bijective surface parameterization for inferring signed
  distance functions from sparse point clouds with grid deformation.
\newblock In {\em Proceedings of the IEEE/CVF Conference on Computer Vision and
  Pattern Recognition (CVPR)}, 2025.

\bibitem{openaisora2024}
{OpenAI}.
\newblock Video generation models as world simulators.
\newblock
  \url{https://openai.com/index/video-generation-models-as-world-simulators/},
  2024.

\bibitem{oquab2023dinov2}
Maxime Oquab, Timoth{\'e}e Darcet, Th{\'e}o Moutakanni, Huy~V. Vo, Marc
  Szafraniec, Vasil Khalidov, Pierre Fernandez, Daniel Haziza, Francisco Massa,
  Alaaeldin El-Nouby, Mahmoud Assran, Nicolas Ballas, Wojciech Galuba, Russell
  Howes, Po-Yao Huang, Shang-Wen Li, Ishan Misra, Michael Rabbat, Vasu Sharma,
  Gabriel Synnaeve, Hu Xu, Herv{\'e} Jegou, Julien Mairal, Patrick Labatut,
  Armand Joulin, and Piotr Bojanowski.
\newblock {DINOv2}: Learning robust visual features without supervision.
\newblock {\em Transactions on Machine Learning Research (TMLR)}, 2024.

\bibitem{pan2025diff4splat}
Panwang Pan, Chenguo Lin, Jingjing Zhao, Chenxin Li, Yuchen Lin, Haopeng Li,
  Honglei Yan, Kairun Wen, Yunlong Lin, Yixuan Yuan, et~al.
\newblock {Diff4Splat}: Controllable {4D} scene generation with latent dynamic
  reconstruction models.
\newblock In {\em Proceedings of the IEEE/CVF Conference on Computer Vision and
  Pattern Recognition (CVPR)}, 2026.

\bibitem{peebles2023scalable}
William Peebles and Saining Xie.
\newblock Scalable diffusion models with transformers.
\newblock In {\em Proceedings of the IEEE/CVF International Conference on
  Computer Vision (ICCV)}, 2023.

\bibitem{Perazzi_CVPR_2016}
Federico Perazzi, Jordi Pont-Tuset, Brian McWilliams, Luc~Van Gool, Markus
  Gross, and Alexander Sorkine-Hornung.
\newblock A benchmark dataset and evaluation methodology for video object
  segmentation.
\newblock In {\em Proceedings of the IEEE/CVF Conference on Computer Vision and
  Pattern Recognition (CVPR)}, 2016.

\bibitem{ren2024l4gm}
Jiawei Ren, Kevin Xie, Ashkan Mirzaei, Hanxue Liang, Xiaohui Zeng, Karsten
  Kreis, Ziwei Liu, Antonio Torralba, Sanja Fidler, Seung~Wook Kim, and Huan
  Ling.
\newblock {L4GM}: Large {4D} {Gaussian} reconstruction model.
\newblock In {\em Advances in Neural Information Processing Systems (NeurIPS)},
  2024.

\bibitem{ren2025gen3c}
Xuanchi Ren, Tianchang Shen, Jiahui Huang, Huan Ling, Yifan Lu, Merlin
  Nimier-David, Thomas M{\"u}ller, Alexander Keller, Sanja Fidler, and Jun Gao.
\newblock {GEN3C}: {3D}-informed world-consistent video generation with precise
  camera control.
\newblock In {\em Proceedings of the IEEE/CVF Conference on Computer Vision and
  Pattern Recognition (CVPR)}, 2025.

\bibitem{shen2026lyra}
Tianchang Shen, Sherwin Bahmani, Kai He, Sangeetha~Grama Srinivasan, Tianshi
  Cao, Jiawei Ren, Ruilong Li, Zian Wang, Nicholas Sharp, Zan Gojcic, et~al.
\newblock Lyra 2.0: Explorable generative {3D} worlds.
\newblock In {\em SIGGRAPH Asia}, 2026.

\bibitem{sun2024splatter}
Yang-Tian Sun, Yi-Hua Huang, Lin Ma, Xiaoyang Lyu, Yan-Pei Cao, and Xiaojuan
  Qi.
\newblock {Splatter a Video}: Video {Gaussian} representation for versatile
  processing.
\newblock In {\em Advances in Neural Information Processing Systems (NeurIPS)},
  2024.

\bibitem{team2025hunyuanworld}
HunyuanWorld Team, Zhenwei Wang, Yuhao Liu, Junta Wu, Zixiao Gu, Haoyuan Wang,
  Xuhui Zuo, Tianyu Huang, Wenhuan Li, Sheng Zhang, et~al.
\newblock {HunyuanWorld} 1.0: Generating immersive, explorable, and interactive
  {3D} worlds from words or pixels.
\newblock {\em arXiv preprint arXiv:2507.21809}, 2025.

\bibitem{team2026inspatio}
InSpatio Team, Donghui Shen, Guofeng Zhang, Haomin Liu, Haoyu Ji, Hujun Bao,
  Hongjia Zhai, Jialin Liu, Jing Guo, Nan Wang, et~al.
\newblock {INSPATIO-WORLD}: A real-time {4D} world simulator via spatiotemporal
  autoregressive modeling.
\newblock {\em arXiv preprint arXiv:2604.07209}, 2026.

\bibitem{team2026advancing}
Robbyant Team, Zelin Gao, Qiuyu Wang, Yanhong Zeng, Jiapeng Zhu, Ka~Leong
  Cheng, Yixuan Li, Hanlin Wang, Yinghao Xu, Shuailei Ma, et~al.
\newblock Advancing open-source world models.
\newblock {\em arXiv preprint arXiv:2601.20540}, 2026.

\bibitem{wan2025wan}
Team Wan, Ang Wang, Baole Ai, Bin Wen, Chaojie Mao, Chen-Wei Xie, Di Chen,
  Feiwu Yu, Haiming Zhao, Jianxiao Yang, et~al.
\newblock Wan: Open and advanced large-scale video generative models.
\newblock {\em arXiv preprint arXiv:2503.20314}, 2025.

\bibitem{wang2025shape}
Qianqian Wang, Vickie Ye, Hang Gao, Weijia Zeng, Jake Austin, Zhengqi Li, and
  Angjoo Kanazawa.
\newblock {Shape of Motion}: {4D} reconstruction from a single video.
\newblock In {\em Proceedings of the IEEE/CVF International Conference on
  Computer Vision (ICCV)}, 2025.

\bibitem{tartanair2020iros}
Wenshan Wang, Delong Zhu, Xiangwei Wang, Yaoyu Hu, Yuheng Qiu, Chen Wang, Yafei
  Hu, Ashish Kapoor, and Sebastian Scherer.
\newblock {TartanAir}: A dataset to push the limits of visual {SLAM}.
\newblock In {\em IEEE/RSJ International Conference on Intelligent Robots and
  Systems (IROS)}, 2020.

\bibitem{wang2025freetimegs}
Yifan Wang, Peishan Yang, Zhen Xu, Jiaming Sun, Zhanhua Zhang, Yong Chen, Hujun
  Bao, Sida Peng, and Xiaowei Zhou.
\newblock {FreeTimeGS}: Free {Gaussian} primitives at anytime anywhere for
  dynamic scene reconstruction.
\newblock In {\em Proceedings of the IEEE/CVF Conference on Computer Vision and
  Pattern Recognition (CVPR)}, 2025.

\bibitem{wang2025pi}
Yifan Wang, Jianjun Zhou, Haoyi Zhu, Wenzheng Chang, Yang Zhou, Zizun Li, Junyi
  Chen, Jiangmiao Pang, Chunhua Shen, and Tong He.
\newblock {$\pi^3$}: Permutation-equivariant visual geometry learning.
\newblock In {\em International Conference on Learning Representations (ICLR)},
  2026.

\bibitem{wang2004image}
Zhou Wang, Alan~C. Bovik, Hamid~R. Sheikh, and Eero~P. Simoncelli.
\newblock Image quality assessment: From error visibility to structural
  similarity.
\newblock {\em IEEE Transactions on Image Processing (TIP)}, 13(4):600--612,
  2004.

\bibitem{wang2026matrix}
Zile Wang, Zexiang Liu, Jiaxing Li, Kaichen Huang, Baixin Xu, Fei Kang, Mengyin
  An, Peiyu Wang, Biao Jiang, Yichen Wei, et~al.
\newblock {Matrix-Game} 3.0: Real-time and streaming interactive world model
  with long-horizon memory.
\newblock {\em arXiv preprint arXiv:2604.08995}, 2026.

\bibitem{worldlabsmarble}
{World Labs}.
\newblock Marble: A multimodal world model.
\newblock \url{https://www.worldlabs.ai/blog/marble-world-model}, 2025.

\bibitem{wu2026streamsplat}
Zike Wu, Qi Yan, Xuanyu Yi, Lele Wang, and Renjie Liao.
\newblock {StreamSplat}: Towards online dynamic {3D} reconstruction from
  uncalibrated video streams.
\newblock In {\em International Conference on Learning Representations (ICLR)},
  2026.

\bibitem{xiang2026vggs}
Peng Xiang, Liang Han, Hui Zhang, Yu-Shen Liu, and Zhizhong Han.
\newblock {VGGS}: {VGGT}-guided {Gaussian} splatting for efficient and faithful
  sparse-view surface reconstruction.
\newblock In {\em Proceedings of the AAAI Conference on Artificial Intelligence
  (AAAI)}, 2026.

\bibitem{xu20254dgt}
Zhen Xu, Zhengqin Li, Zhao Dong, Xiaowei Zhou, Richard Newcombe, and Zhaoyang
  Lv.
\newblock {4DGT}: Learning a {4D} {Gaussian} transformer using real-world
  monocular videos.
\newblock In {\em Advances in Neural Information Processing Systems (NeurIPS)},
  2025.

\bibitem{yang2024depthv2}
Lihe Yang, Bingyi Kang, Zilong Huang, Zhen Zhao, Xiaogang Xu, Jiashi Feng, and
  Hengshuang Zhao.
\newblock {Depth Anything V2}.
\newblock In {\em Advances in Neural Information Processing Systems (NeurIPS)},
  2024.

\bibitem{yang2026neoverse}
Yuxue Yang, Lue Fan, Ziqi Shi, Junran Peng, Feng Wang, and Zhaoxiang Zhang.
\newblock {NeoVerse}: Enhancing {4D} world model with in-the-wild monocular
  videos.
\newblock In {\em Proceedings of the IEEE/CVF Conference on Computer Vision and
  Pattern Recognition (CVPR)}, 2026.

\bibitem{yang2025matrix}
Zhongqi Yang, Wenhang Ge, Yuqi Li, Jiaqi Chen, Haoyuan Li, Mengyin An, Fei
  Kang, Hua Xue, Baixin Xu, Yuyang Yin, et~al.
\newblock {Matrix-3D}: Omnidirectional explorable {3D} world generation.
\newblock {\em arXiv preprint arXiv:2508.08086}, 2025.

\bibitem{cogvideox}
Zhuoyi Yang, Jiayan Teng, Wendi Zheng, Ming Ding, Shiyu Huang, Jiazheng Xu,
  Yuanming Yang, Wenyi Hong, Xiaohan Zhang, Guanyu Feng, et~al.
\newblock {CogVideoX}: Text-to-video diffusion models with an expert
  transformer.
\newblock In {\em International Conference on Learning Representations (ICLR)},
  2025.

\bibitem{yangreal}
Zeyu Yang, Hongye Yang, Zijie Pan, and Li Zhang.
\newblock Real-time photorealistic dynamic scene representation and rendering
  with {4D} {Gaussian} splatting.
\newblock In {\em International Conference on Learning Representations (ICLR)},
  2024.

\bibitem{Yu_2025_ICCV}
Mark Yu, Wenbo Hu, Jinbo Xing, and Ying Shan.
\newblock {TrajectoryCrafter}: Redirecting camera trajectory for monocular
  videos via diffusion models.
\newblock In {\em Proceedings of the IEEE/CVF International Conference on
  Computer Vision (ICCV)}, 2025.

\bibitem{zhang2018unreasonable}
Richard Zhang, Phillip Isola, Alexei~A. Efros, Eli Shechtman, and Oliver Wang.
\newblock The unreasonable effectiveness of deep features as a perceptual
  metric.
\newblock In {\em Proceedings of the IEEE/CVF Conference on Computer Vision and
  Pattern Recognition (CVPR)}, 2018.

\bibitem{zhang2025materialrefgs}
Wenyuan Zhang, Jimin Tang, Weiqi Zhang, Yi Fang, Yu-Shen Liu, and Zhizhong Han.
\newblock {MaterialRefGS}: Reflective {Gaussian} splatting with multi-view
  consistent material inference.
\newblock In {\em Advances in Neural Information Processing Systems (NeurIPS)},
  2025.

\bibitem{zhang2026tesla}
Weiqi Zhang, Junsheng Zhou, Juntong Fang, Kanle Shi, Shenkun Xu, and Yu-Shen
  Liu.
\newblock {TESLA}: Native {4D} {Gaussian} splatting generation with temporally
  structured latents.
\newblock In {\em Advances in Neural Information Processing Systems (NeurIPS)},
  2026.

\bibitem{zhang2026gaussiangrow}
Weiqi Zhang, Junsheng Zhou, Haotian Geng, Kanle Shi, Shenkun Xu, Yi Fang, and
  Yu-Shen Liu.
\newblock {GaussianGrow}: Geometry-aware {Gaussian} growing from {3D} point
  clouds with text guidance.
\newblock In {\em Proceedings of the IEEE/CVF Conference on Computer Vision and
  Pattern Recognition (CVPR)}, 2026.

\bibitem{zhang2025gap}
Weiqi Zhang, Junsheng Zhou, Haotian Geng, Wenyuan Zhang, and Yu-Shen Liu.
\newblock {GAP}: {Gaussianize} any point clouds with text guidance.
\newblock In {\em Proceedings of the IEEE/CVF International Conference on
  Computer Vision (ICCV)}, 2025.

\bibitem{zhang2026gamma}
Weiqi Zhang, Junsheng Zhou, Xuancheng Zhang, Juntong Fang, Zequn Chen, Donglin
  Di, and Yu-Shen Liu.
\newblock {GAMMA}: Scalable {4D} {Gaussian} reconstruction model for novel view
  synthesis of monocular videos.
\newblock In {\em Advances in Neural Information Processing Systems (NeurIPS)},
  2026.

\bibitem{zheng2023point}
Yang Zheng, Adam~W. Harley, Bokui Shen, Gordon Wetzstein, and Leonidas~J.
  Guibas.
\newblock {PointOdyssey}: A large-scale synthetic dataset for long-term point
  tracking.
\newblock In {\em Proceedings of the IEEE/CVF International Conference on
  Computer Vision (ICCV)}, 2023.

\bibitem{zhou2024diffgs}
Junsheng Zhou, Weiqi Zhang, and Yu-Shen Liu.
\newblock {DiffGS}: Functional {Gaussian} splatting diffusion.
\newblock In {\em Advances in Neural Information Processing Systems (NeurIPS)},
  2024.

\bibitem{zhou2024udiff}
Junsheng Zhou, Weiqi Zhang, Baorui Ma, Kanle Shi, Yu-Shen Liu, and Zhizhong
  Han.
\newblock {UDiFF}: Generating conditional unsigned distance fields with optimal
  wavelet diffusion.
\newblock In {\em Proceedings of the IEEE/CVF Conference on Computer Vision and
  Pattern Recognition (CVPR)}, 2024.

\bibitem{zhou2026udfstudio}
Junsheng Zhou, Weiqi Zhang, Baorui Ma, Kanle Shi, Yu-Shen Liu, and Zhizhong
  Han.
\newblock {UDFStudio}: A unified framework of datasets, benchmarks and
  generative models for unsigned distance functions.
\newblock {\em IEEE Transactions on Pattern Analysis and Machine Intelligence
  (TPAMI)}, 2026.

\bibitem{zhou2025omniworld}
Yang Zhou, Yifan Wang, Jianjun Zhou, Wenzheng Chang, Haoyu Guo, Zizun Li,
  Kaijing Ma, Xinyue Li, Yating Wang, Haoyi Zhu, Mingyu Liu, Dingning Liu,
  Jiange Yang, Zhoujie Fu, Junyi Chen, Chunhua Shen, Jiangmiao Pang, Kaipeng
  Zhang, and Tong He.
\newblock {OmniWorld}: A multi-domain and multi-modal dataset for {4D} world
  modeling.
\newblock In {\em International Conference on Learning Representations (ICLR)},
  2026.

\end{thebibliography}

\newpage
\appendix

\section{Implementation Details}
\label{sec:impl}

The 4DRM encoder is built on DINOv2-Large~\cite{oquab2023dinov2} with 24 transformer blocks of dimension 1024. The 4DRM is initialized from the pretrained MapAnything weights~\cite{keetha2025mapanything}.
The domain adapter is injected at layer $\ell{=}4$, retaining the upper 20 pretrained transformer blocks as a geometric foundation for the 4DRM. 
The Linear branch projects the 16-channel Wan video latent to 1024 channels, initialized via closed-form ridge regression against the fourth-layer DINOv2 activations. The Perceiver Bridge employs 4 cross-attention blocks with 16 heads and 1024 learnable queries, and a temporal interleave of factor 4 inverts the video VAE's temporal compression. All three stages are optimized with AdamW using a weight decay of $10^{-4}$, a cosine schedule with 2\,000 linear warmup steps, a base learning rate of $5{\times}10^{-6}$, and gradient clipping at a global norm of 1.0. 

Stage~1 trains the full 4DRM for 20 epochs on video clips with ground-truth camera poses and depth annotations. The appearance loss weights are $\alpha{=}2.0$ and $\beta{=}0.2$, the geometry weights are $\mu{=}1.0$ and $\nu{=}0.5$, and the dynamic regularization coefficients are $\eta_1{=}\eta_2{=}\eta_3{=}0.3$. Stage~2 freezes the Stage-1 model as the teacher and trains the domain adapter and the student 4DRM for 30 additional epochs with a $10{\times}$ learning rate multiplier on the adapter parameters. Since the frozen teacher provides all supervisory signals through distillation, Stage~2 requires only raw RGB videos as input, removing the dependency on pose and depth annotations and enabling training on large-scale unannotated data. The feature alignment weight $\lambda$ decays from 1.0 to 0 following a cosine schedule. All experiments are conducted on 16 NVIDIA H200 GPUs. The Stage-1 training corpus combines DynamicReplica~\cite{karaev2023dynamicstereo}, SAILVOS3D~\cite{HuCVPR2021}, Spring~\cite{Mehl2023_Spring}, TartanAir~\cite{tartanair2020iros}, PointOdyssey~\cite{zheng2023point}, and OmniWorld~\cite{zhou2025omniworld}, each providing ground-truth camera poses and depth. Because these datasets differ in size by orders of magnitude, we resample the smaller ones by the factors listed in Table~\ref{tab:datasets}, so that every source contributes to geometric supervision. In Stage~2, we further incorporate Panda-70M~\cite{chen2024panda70m} to expand the training distribution beyond annotated domains; it supplies real-world appearance diversity without geometric labels. Stage~3 continues to train the domain adapter and student 4DRM for 10 additional epochs with the end-to-end rendering loss, using the same optimizer configuration as Stage~2. The distillation regularization weight $\lambda_{\text{reg}}$ is initialized to 0.5 and linearly decayed to zero over the course of training.

\begin{table}[htbp]
\centering
\caption{Training datasets. Resample is the factor by which each dataset is repeated during training.}
\label{tab:datasets}
\setlength{\tabcolsep}{5pt}
\begin{tabular*}{\linewidth}{@{\extracolsep{\fill}}lccccrr@{}}
\toprule
Dataset & Dynamic & Pose & Depth & Real & \#Scenes & Resample \\
\midrule
PointOdyssey~\cite{zheng2023point} & \ding{51} & \ding{51} & \ding{51} & \ding{55} & 0.1K & 100$\times$ \\
TartanAir~\cite{tartanair2020iros} & \ding{51} & \ding{51} & \ding{51} & \ding{55} & 0.5K & 50$\times$ \\
DynamicReplica~\cite{karaev2023dynamicstereo} & \ding{51} & \ding{51} & \ding{51} & \ding{55} & 0.5K & 20$\times$ \\
OmniWorld-Game~\cite{zhou2025omniworld} & \ding{51} & \ding{51} & \ding{51} & \ding{55} & 0.06K & 100$\times$ \\
SAIL-VOS 3D~\cite{HuCVPR2021} & \ding{51} & \ding{51} & \ding{51} & \ding{55} & 0.16K & 50$\times$ \\
Spring~\cite{Mehl2023_Spring} & \ding{51} & \ding{51} & \ding{51} & \ding{55} & 0.03K & 1000$\times$ \\
Panda-70M~\cite{chen2024panda70m} (subset) & \ding{51} & \ding{55} & \ding{55} & \ding{51} & 2.4M & 1$\times$ \\
\bottomrule
\end{tabular*}
\end{table}

\section{Applications}
\label{sec:applications}

\paragraph{Text/Image-to-world generation.}
As illustrated in Fig.~\ref{fig:teaser}, AEON synthesizes high-quality 4D Gaussian Splatting (4DGS) environments from text or image prompts. This is achieved by first generating video latents using Wan-DiT conditioned on the input prompts. These latents are subsequently translated into 4D spatial-temporal latents and decoded to reconstruct the final 4DGS world.

\paragraph{Video-to-world reconstruction.}
For video-to-world reconstruction, input videos are encoded into video latents using Wan-VAE. These latents are then processed to generate 4DGS scenes, following the same pipeline as our text/image-to-world generation.

\paragraph{Novel view synthesis.}
The generated 4DGS representations inherently support real-time novel view synthesis across arbitrary viewpoints and timesteps.

\paragraph{Physical status estimation.}
The generated 4DGS encapsulates rich geometric and physical properties, which can be directly extracted to derive depth maps, normal maps, motion masks, and optical flow. Specifically, motion masks are obtained by filtering the lifespans of the Gaussians, as dynamic elements typically correspond to Gaussians with shorter lifespans.

\paragraph{3D world generation.}
Beyond dynamic scene generation, AEON is also highly effective for static 3D world generation. An example of this capability is provided in the third row of Fig.~\ref{fig:teaser}.

\paragraph{Scene editing and simulation.}
Because AEON generates explicit 4D representations, it inherently facilitates seamless scene editing and structural decomposition for downstream simulations (shown in the bottom row of Fig.~\ref{fig:teaser}). By leveraging the motion masks derived from Gaussian lifespan filtering, we can cleanly separate a generated scene into its static and dynamic components. These isolated parts from various generated 4DGS worlds can then be freely recomposed to simulate novel scenarios. The same explicit structure is what embodied simulation needs. The metric geometry of the static component can supply occupancy and free space for an agent, the dynamic primitives move according to their own velocities, and real-time rasterization renders egocentric observations at every step. An agent can therefore be placed in a generated world, roll out its actions, and receive rendered observations in a closed loop, and because the world is persistent, the same scenario can be replayed for repeatable evaluation.

\section{Evolution of 4DGS Representation}
\label{sec:evolution}

\paragraph{Temporal opacity evolution.}
To faithfully represent the appearance and disappearance of scene elements over time, the opacity of each primitive is modulated by a Gaussian kernel centered at its reference timestep $t_i$. Given a query time $t$, the effective opacity is computed as
\begin{equation}
\sigma_i=\sqrt{\frac{(l_i/2)^2}{-2\log(\epsilon)}},\qquad o_i(t)=o_i \cdot \exp\!\left(-\frac{(t-t_i)^2}{2\sigma_i^{2}}\right),
\end{equation}
where $\epsilon = 0.05$ serves as the activation threshold at the duration boundary. This temporal gating mechanism ensures full visibility near the reference timestep while producing smooth fade-in and fade-out transitions. As a result, the representation can naturally accommodate transient phenomena such as fire, splashes, and object entrances without resorting to explicit per-object tracking.
 
\paragraph{Velocity-driven spatial evolution.}
The position and orientation of each primitive evolve over time according to a constant-velocity motion model. At query time $t$, the updated position and rotation are given by
\begin{equation}
\mathbf{x}_i(t) = \mathbf{x}_i + \mathbf{v}_i \cdot (t - t_i),
\end{equation}
\begin{equation}
\mathbf{r}_i(t) = \mathbf{r}_i \otimes \mathrm{AxisAngle2Quat}(\boldsymbol{\omega}_i \cdot (t - t_i)),
\end{equation}
where $\otimes$ denotes quaternion multiplication. The scaling $\mathbf{s}_i$ remains fixed across time, ensuring that each primitive preserves its intrinsic shape throughout the sequence. Factoring translational and rotational dynamics into independent velocity terms yields a compact yet expressive motion parameterization that supports rigid-body motion while incurring minimal per-primitive overhead. Combined with the temporal opacity mechanism above, the full 4DGS representation enables AEON to unify the photorealistic rendering capability of video models with the geometric consistency and persistence of 3D models.

\section{Evaluation Metrics}
\label{sec:metrics}

We describe all evaluation metrics used in the main paper. Our experiments rely on two families of metrics: per-frame reconstruction metrics that measure pixel-level fidelity and VBench metrics that assess perceptual video quality along multiple complementary dimensions.

\paragraph{Reconstruction metrics.}
We evaluate dynamic scene reconstruction with three standard metrics that operate at increasing levels of perceptual abstraction. Peak Signal-to-Noise Ratio (PSNR) measures per-pixel reconstruction accuracy in decibels:
\begin{equation}
\text{PSNR} = 10 \cdot \log_{10} \frac{R^2}{\frac{1}{N}\sum_{i=1}^{N}(\mathbf{I}_i - \tilde{\mathbf{I}}_i)^2},
\end{equation}
where $R$ is the maximum pixel intensity, $N$ is the total number of pixels, and $\mathbf{I}$, $\tilde{\mathbf{I}}$ denote the ground-truth and rendered frames respectively. Higher values indicate less distortion. Structural Similarity Index (SSIM)~\cite{wang2004image} moves beyond pixel-level comparison by jointly evaluating local luminance, contrast, and structural patterns:
\begin{equation}
\text{SSIM}(\mathbf{I}, \tilde{\mathbf{I}}) = \frac{(2\mu_{\mathbf{I}}\mu_{\tilde{\mathbf{I}}} + C_1)(2\sigma_{\mathbf{I}\tilde{\mathbf{I}}} + C_2)}{(\mu_{\mathbf{I}}^2 + \mu_{\tilde{\mathbf{I}}}^2 + C_1)(\sigma_{\mathbf{I}}^2 + \sigma_{\tilde{\mathbf{I}}}^2 + C_2)},
\end{equation}
where $\mu$ and $\sigma$ denote the local mean and standard deviation computed over sliding windows, $\sigma_{\mathbf{I}\tilde{\mathbf{I}}}$ is the local cross-covariance, and $C_1$, $C_2$ are small stability constants. SSIM ranges from 0 to 1, with higher values indicating greater structural agreement. Learned Perceptual Image Patch Similarity (LPIPS)~\cite{zhang2018unreasonable} further captures high-level perceptual differences by comparing deep features extracted from a pretrained network:
\begin{equation}
\text{LPIPS}(\mathbf{I}, \tilde{\mathbf{I}}) = \sum_{l} w_l \left\| \phi_l(\mathbf{I}) - \phi_l(\tilde{\mathbf{I}}) \right\|_2^2,
\end{equation}
where $\phi_l$ denotes the normalized activations at layer $l$ and $w_l$ are learned linear weights. Lower LPIPS values indicate higher perceptual similarity.

\paragraph{VBench video quality metrics.}
For 4D generation and novel trajectory rendering we adopt six metrics from the VBench~\cite{huang2023vbench} benchmark, each targeting a distinct aspect of video quality. \textbf{I2V Subject} measures the visual fidelity between the conditioning image and the generated subject, quantifying how faithfully the output preserves the identity of the input. \textbf{Subject Consistency} evaluates the temporal coherence of the foreground subject across frames by computing pairwise feature similarity over the video. \textbf{Background Consistency} similarly measures the stability of background regions to detect unwanted scene drift. \textbf{Temporal Flickering} captures high-frequency luminance and color variations between consecutive frames, where higher scores correspond to smoother transitions. \textbf{Motion Smoothness} assesses whether object trajectories follow physically plausible paths without abrupt jumps. \textbf{Aesthetic Quality} and \textbf{Imaging Quality} respectively evaluate the overall visual appeal and the low-level image attributes such as sharpness and color fidelity. All VBench metrics are reported on a scale where higher values indicate better performance.

\section{Additional Ablation Studies}
\label{sec:ablation_details}

Table~\ref{tab:ablation_more} reports additional ablations on Dycheck and DaVis, where each variant changes one component of the full model.

\begin{table}[htbp]
    \centering
    \caption{Additional ablation studies on Dycheck and DaVis.}
    \label{tab:ablation_more}
    \setlength{\tabcolsep}{6pt}
    \resizebox{\linewidth}{!}{
        \begin{tabular}{llcccccc}
        \toprule
        & \multirow{2}{*}{Variant}
        & \multicolumn{3}{c}{Dycheck}
        & \multicolumn{3}{c}{DaVis} \\
        \cmidrule(lr){3-5} \cmidrule(lr){6-8}
        & & PSNR $\uparrow$ & SSIM $\uparrow$ & LPIPS $\downarrow$
        & PSNR $\uparrow$ & SSIM $\uparrow$ & LPIPS $\downarrow$ \\
        \midrule
        \multirow{3}{*}{Adapter} & Conv3D projection & 12.086 & 0.314 & 0.606 & 12.813 & 0.276 & 0.701 \\
        & Conv branch only      & 12.332 & 0.333 & 0.533 & 15.409 & 0.465 & 0.533 \\
        & Perceiver branch only & 13.086 & 0.389 & 0.596 & 14.262 & 0.442 & 0.552 \\
        \midrule
        Training & W/o Stage 3 & 13.670 & 0.361 & 0.517 & 22.953 & 0.705 & 0.299 \\
        \midrule
        \multirow{2}{*}{4DRM} & W/o geometric prior & 12.341 & 0.358 & 0.543 & 22.268 & 0.695 & 0.349 \\
        & W/o $\mathcal{L}_{\text{geo}}$ & 12.260 & 0.331 & 0.578 & 20.347 & 0.655 & 0.323 \\
        \midrule
        \multicolumn{2}{l}{Full} & \textbf{14.487} & \textbf{0.421} & \textbf{0.469} & \textbf{24.661} & \textbf{0.771} & \textbf{0.243} \\
        \bottomrule
        \end{tabular}
    }
\end{table}

\paragraph{Adapter.} \textit{Conv3D projection} replaces the dual-path adapter with a single Conv3D projection, while \textit{Conv branch only} and \textit{Perceiver branch only} use only a convolutional branch or only the Perceiver branch, respectively. All three variants lower DaVis PSNR to 12.8--15.4\,dB, so no single path can bridge the two latent spaces on its own.

\paragraph{Training.} \textit{W/o Stage~3} stops after Stage~2 and keeps Stages~1 and~2 unchanged. Skipping Stage~3 reduces DaVis PSNR by 1.7\,dB, demonstrating the benefit of end-to-end rendering refinement.

\paragraph{4DRM.} \textit{W/o geometric prior} trains 4DRM without the MapAnything initialization, and \textit{W/o $\mathcal{L}_{\text{geo}}$} removes the depth and normal supervision of Stage~1. Both variants perform worse than the full model on Dycheck and DaVis, so the quality of 4DRM matters, although performance degrades gracefully rather than collapsing.

\section{Additional Qualitative Results}

\begin{figure}[htbp]
    \centering
    \includegraphics[width=\linewidth]{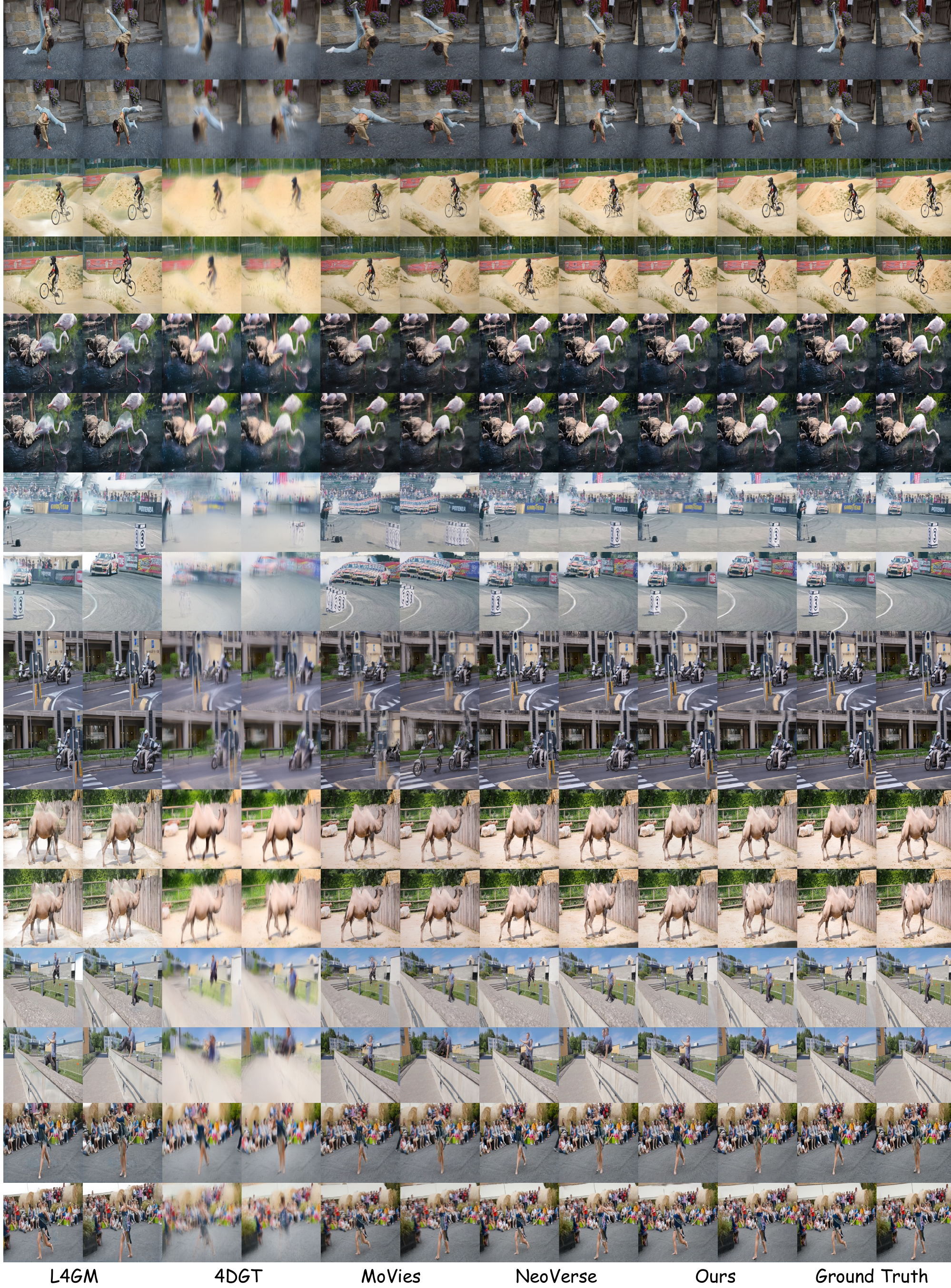}
    \caption{Additional qualitative reconstruction comparisons on DaVis. AEON produces more coherent geometry and finer details in dynamic regions.}
    \label{fig:reconstruction_2}
\end{figure}

We provide additional qualitative reconstruction comparisons on the DaVis benchmark in Fig.~\ref{fig:reconstruction_2}. These results further corroborate that AEON recovers sharper appearance details and more temporally coherent geometry than baselines, particularly in regions with fast motion and transient occlusions.

\end{document}